%% file: conference_101719.tex
\documentclass[conference]{IEEEtran}
\IEEEoverridecommandlockouts
\usepackage{cite}
\usepackage{amsmath,amssymb,amsfonts}
\usepackage{algorithmic}
\usepackage{graphicx}
\usepackage{textcomp}
\usepackage{xcolor}
\usepackage[mode=text]{siunitx} 
\usepackage{tikz} 
\usetikzlibrary{positioning} 
\usepackage[inkscapelatex=false]{svg} 
\usepackage{algorithm}    
\usepackage{booktabs}     
\usepackage[mode=text]{siunitx} 
\usepackage{fancyhdr}
\usepackage{hyperref}
\hypersetup{
    colorlinks,
    linkcolor={red!50!black},
    citecolor={blue!50!black},
    urlcolor={blue!80!black}
}

\usepackage{tikz} 
\usetikzlibrary{shapes.geometric, arrows.meta, positioning}

\tikzset{
  io/.style={
    trapezium, trapezium left angle=75, trapezium right angle=105,
    draw, fill=green!40, text width=2.2cm, align=center, minimum height=0.7cm
  },
  process/.style={
    rectangle, draw, fill=blue!25,
    text width=2.8cm, align=center, minimum height=0.8cm
  },
  decision/.style={
    diamond, draw, fill=yellow!50, aspect=2,
    text width=2.2cm, align=center, inner sep=1pt
  },
  adjust/.style={
    rectangle, draw, fill=orange!40,
    text width=2cm, align=center, minimum height=0.8cm
  },
  pipeline/.style={
    rectangle, draw, fill=violet!25,
    text width=2.8cm, align=center, minimum height=0.8cm
  },
  arr/.style={-{Stealth}, thick},
}

\def\BibTeX{{\rm B\kern-.05em{\sc i\kern-.025em b}\kern-.08em
    T\kern-.1667em\lower.7ex\hbox{E}\kern-.125emX}}
\begin{document}

\newif\ifdraftcolor
\draftcolortrue 
\ifdraftcolor
\newcommand{\todo}[1]{{\color{red}TODO: #1}}
\newcommand{\fclad}[1]{{\color{olive}(fclad) #1}}
\else
\newcommand{\todo}[1]{}
\newcommand{\fclad}[1]{}
\fi

\title{EventShiftFlow: Towards Hardware-efficient FPGA-based Flow Estimation\\
\thanks{All authors are with the University of Pennsylvania, Philadelphia, PA.\\
Corresponding author: \texttt{alari@seas.upenn.edu}.}
}

\author{Arianna Alonso Bizzi, Fernando Cladera, and C. J. Taylor}

\maketitle
\fancypagestyle{withfooter}{
\renewcommand{\headrulewidth}{0pt}
\fancyfoot[C]{\footnotesize Accepted to the Challenges and Opportunities of Neuromorphic Field Robotics and Automation IEEE ICRA Workshop - 2026}
}
\thispagestyle{withfooter}
\pagestyle{withfooter}
\begin{abstract}
Event-based vision sensors offer asynchronous, high-temporal-resolution measurements that are attractive for low-latency robotic perception, but many event-based motion estimation methods are computationally intensive and difficult to map to FPGA hardware. We present a streaming velocity estimator that discretizes asynchronous events into fixed-duration time bins, constructs a 1-bit spatial occupancy grid, and evaluates multiple velocity hypotheses in parallel using only fixed-width integer logic---shift registers, counters, comparators, and small LUT-mapped multiplies---with no dividers and no DSP blocks. It requires no frame reconstruction, no floating-point arithmetic, and no iterative optimization. The method deliberately trades dense sub-pixel optical flow for a sparse, quantized velocity estimate at each active pixel, suited to low-latency tasks such as reactive obstacle avoidance on size-, weight-, and power-constrained platforms. On noisy synthetic data with known ground-truth velocities, the method recovers both magnitude and direction, with magnitude estimates being most challenged when objects of different velocities intersect. On a real event-camera sequence, directional accuracy reaches 99.5\% across all four evaluated motion segments, with performance remaining robust across occupancy densities in the 10--40\% range. 
We characterize the algorithm's density-dependent behavior, present a parameter sensitivity analysis, show that the proposed datapath requires less than 2 kB of storage, and implement a single-axis prototype on a low-cost Xilinx Artix-7.


Project page: \url{https://alonsobizzi.github.io/eventshiftflow-site/}.
\end{abstract}


\input{Sections/intro_literature}
\input{Sections/setup_methods}

\input{Sections/hardware}

\input{Sections/results}
\input{Sections/outlook}




\bibliographystyle{IEEEtran}
\bibliography{references}


\end{document}

%% file: Sections/intro_literature.tex
\section{Introduction}
Event cameras are bio-inspired sensors that asynchronously report per-pixel brightness changes with microsecond temporal resolution, high dynamic range, and low power consumption \cite{gallego2022survey}. These properties make them attractive for low-latency motion estimation in robotic systems, particularly of size, weight, and power (SWaP) constrained platforms such as micro aerial vehicles. However, most existing methods for estimating motion from events, whether based on local plane fitting\cite{benosman2014event}, contrast maximization \cite{gallego2018contrast}, or deep learning \cite{zhu2018evflownet, gehrig2021eraft}, require operations that are expensive in hardware: floating-point arithmetic, iterative optimization, or large memory buffers for frame reconstruction.

Field-programmable gate arrays (FPGAs) and application-specific integrated circuits (ASICs) are natural targets for event-driven processing because they can handle each incoming event with fixed, sub-microsecond latency in a dedicated datapath, matching the asynchronous, low-latency nature of the sensor itself. Unlike CPUs or GPUs, which batch events for processing, reconfigurable hardware can feature parallel processing lanes that operate continuously on the event stream with deterministic timing and lower power consumption. This makes FPGA-based processing particularly attractive for resource-constrained robotic platforms where latency, power, and weight budgets preclude general-purpose processors. FPGAs are preferred over ASICs for rapid design iteration. They allow in-field reconfiguration as the algorithm matures, without the high non-recurring engineering costs of custom silicon.

For deployment on FPGA or ASIC, the algorithm must map to a fixed datapath with predictable latency, bounded memory and thus no unbounded, iterative loops. Existing methods do not fit this constraint: plane-fitting methods require least-squares solves per event; contrast maximization requires a search over a continuous parameter space; and learned methods require networks with tens of millions of parameters and corresponding multiply-accumulate operations per inference \cite{zhu2018evflownet, gehrig2021eraft}. FPGA implementations exist, but many remain resource-intensive. 

\begin{figure}[t]
\centerline{\includegraphics[width=0.999\columnwidth]{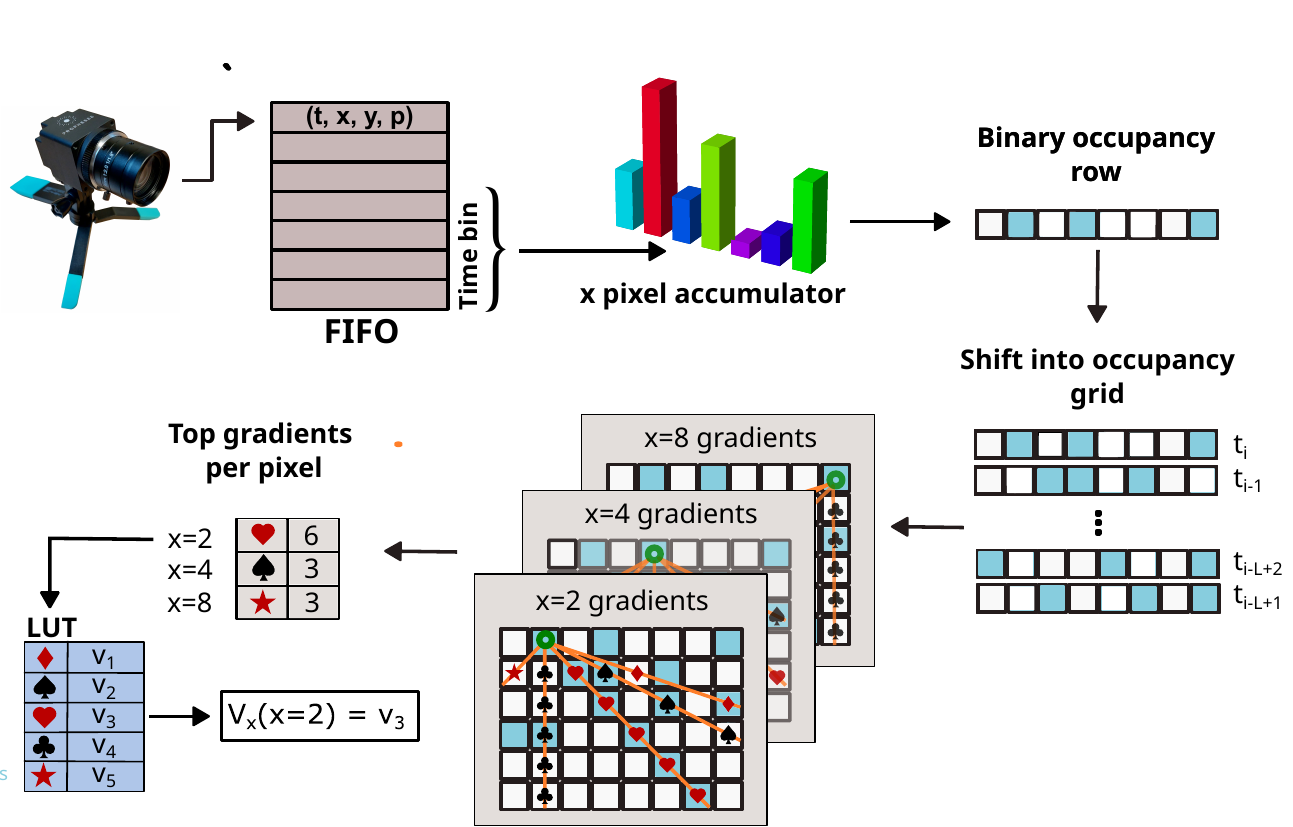}}
\caption{Bitvector-based motion estimation pipeline. Event data for each axis is accumulated and thresholded to be discretized into binary occupancy representations. A spatiotemporal occupancy grid is constructed by keeping a buffer of past occupancy representations. Motion is inferred via parallel evaluation of delay hypotheses using lightweight bitvector scoring, enabling efficient, parallelizable hardware implementation. The figure represents the x motion pipeline, but the same approach is applied in parallel to the y axis.}
\label{fig:flowchart}
\end{figure}

We propose a different approach that pushes the limit of how much we can shrink and simplify the hardware whilst still recovering meaningful flow information in real-time.
In its current form, the algorithm requires no frame reconstruction, no floating-point arithmetic, and no iterative optimization. The entire state fits in under 2kB of on-chip storage for a DAVIS240C sensor. We validate the method on noisy, synthetic data with known ground-truth velocities and on real event camera recordings from the RPG dataset \cite{mueggler2017dataset}, where we characterize a density-dependent directional bias and analyze parameter sensitivity. We discuss the direct mapping to an FPGA datapath and provide resource and latency estimates.

In summary, the contributions of this paper are:
\begin{itemize}
    \item \textbf{An FPGA-aware bitvector-based motion estimation algorithm}, with low-resource utilization yet acceptable performance for normal-flow estimation operations.
    \item \textbf{Evaluation of this algorithm} in synthetic and real-world data.
    \item A \textbf{resource and latency characterization} of the proposed method, targeting low-resource FPGAs.
\end{itemize}

An overview of our algorithm is seen in Fig.~\ref{fig:flowchart}.

\section{Related Work}

\subsection{Event-Based Optical Flow}

The problem of estimating visual motion from event streams has been approached from several directions. Benosman et al.~\cite{benosman2014event} introduced local plane fitting on the surface of active events, estimating optical flow from the slope of a fitted plane in $(x, y, t)$ space. While elegant and per-event, the method requires a least-squares solve per event and is sensitive to noise, requiring outlier rejection schemes~\cite{rueckauer2016evaluation}. 
Gallego et al.~\cite{gallego2018contrast} proposed contrast maximization, a framework that searches for the motion parameters that best align events when warped along candidate trajectories. This produces accurate flow but requires iterative optimization over a continuous parameter space, making real-time hardware implementation challenging.

Learning-based methods have achieved the lowest errors on standard benchmarks. EV-FlowNet~\cite{zhu2018evflownet} and E-RAFT~\cite{gehrig2021eraft} use convolutional architectures to predict dense optical flow from event representations, with E-RAFT achieving state-of-the-art accuracy on the DSEC benchmark. Schnider et al.~\cite{schnider2023neuromorphic} demonstrated a spiking neural network variant with reduced complexity for real-time deployment. However, even reduced models require millions of parameters and dedicated neural network accelerators.

\subsection{Hardware Implementations}

Liu and Delbruck developed a series of FPGA optical flow implementations for event cameras~\cite{liu2017blockmatch, liu2019abmof}, culminating in EDFLOW~\cite{liu2022edflow}, which combines adaptive block matching for optical flow extraction (ABMOF) with keypoint detection (SFAST) on a custom DVS+FPGA platform. EDFLOW achieves accuracy comparable to EV-FlowNet while consuming roughly 100$\times$ less power, processing block matches at 123\,GOp/s. However, the design requires a 25$\times$25 pixel block match per keypoint and a powerful Xilinx Zynq XC7Z100 SoC. ABMOF requires four full-density image buffers with 4 bits per pixel, so the buffer scales as $4\text{ bits}\times 4\text{ slices}\times x\times y$, and implementing this with SFAST requires an additional 12 slices at lower resolution, giving a size of 247kB for the combined memory slices.

Plane-fitting algorithms have also been attempted on FPGA. Aung et al.~\cite{aung2018fpga} implemented Benosman's plane-fitting algorithm, achieving sub-microsecond latency with a fully pipelined design capable of 100 million plane fits per second, but requiring division and square-root pipelines for the least-squares computation. Haessig et al.~\cite{haessig2018spiking} implemented a spiking optical flow estimator on IBM's TrueNorth neuromorphic chip, achieving low power consumption but limited to sparse flow estimates. A recent survey by Kowalczyk et al.~\cite{kowalczyk2024fpga} reviews 60 FPGA implementations for event data processing, noting that optical flow remains one of the more computationally demanding tasks.

\subsection{Positioning}

Our approach trades per-pixel flow accuracy for extreme hardware simplicity and low memory requirements. Rather than computing dense optical flow, we detect the dominant velocity at each active pixel using a discrete hypothesis search, scored by counting 1-bit coincidences along diagonal traces in a shift-register occupancy grid. The entire computation uses only fixed-width integer arithmetic implemented with registers, counters, comparators and small LUT-based logic.


This places our method at the minimal-complexity end of the design space, suitable as a lightweight front-end for miniaturized neuromorphic robotic systems where approximate motion cues that allow data sparsification at low latency are more valuable than precise flow at higher cost.

%% file: Sections/setup_methods.tex
\section{Experimental Setup}

\subsection{Prototyping}

The algorithm was prototyped and validated in Python before considering hardware mapping. This allowed rapid iteration on the scoring function, parameter tuning, and diagnostic instrumentation that would be impractical in RTL. The Python prototype simulates data ingestion in the same streaming order they would arrive in hardware: one event time bin at a time, with no lookahead or batch preprocessing. 

\subsection{Synthetic Data}
\label{sec:synth_data}

To validate the algorithm in a controlled setting, we generated noisy, synthetic event streams with known ground-truth velocities. The simulator places multiple objects of configurable width, orientation, and velocity on a 240$\times$180 pixel sensor. At each asynchronous microsecond timestamp, each object advances by its velocity vector, and events are generated at pixels where the object's leading or trailing pixels cross. Timestamp jitter and Gaussian noise events are added uniformly across the sensor at a configurable rate to test robustness.

The base synthetic dataset used for evaluation consists of multiple bars moving in arbitrary directions, with velocities spanning the range of $\pm$0.03\,px/\unit{\micro \second} with a discretization of 0.005\,px/\unit{\micro \second} and 5\% event-rate noise.

Our synthetic data generator was extended to support arbitrary, bitmap-defined shapes with arbitrary velocities as well as angular and translational accelerations. This allows for grounded testing of more realistic motion and provides a transition route to the RPG Event Camera Dataset \cite{mueggler2017dataset}. Examples of the synthetic data can be observed in the first five rows of Fig. \ref{fig:frames}.

\subsection{Real Data}
\label{sec:real_data}

For real-world evaluation, we use the \textit{shapes\_rotation} sequence from the RPG Event Camera Dataset \cite{mueggler2017dataset}, recorded with a DAVIS240C sensor (240$\times$180 pixels). The sequence contains several geometric shapes (triangles, squares, circles) undergoing 6-DoF handheld camera motion over approximately 7 seconds. Events are provided as $(t, x, y, p)$ tuples with timestamps in seconds, which we convert to microseconds for processing.

\subsection{Ground Truth Extraction}
\label{sec:gt}

Obtaining dense ground-truth optical flow for event camera data is an open challenge. Our synthetic data allowed precise ground truth tracking. For the real camera data, we employ two complementary approaches:

\subsubsection{Manual tracking} Initially, we manually tracked identifiable features (corners and edges of geometric shapes) across reconstructed event video frames, recording pixel positions at regular time intervals. From these trajectories, we extracted piecewise-constant velocity segments with direction and approximate magnitude. While coarse, this provides reliable directional ground truth for evaluating whether the algorithm correctly identifies the sign of motion. 


\subsubsection{Frame-based optical flow} To obtain denser ground-truth velocity estimates, we reconstruct intensity frames from the event stream using E2VID \cite{rebecq2019e2vid}, a recurrent neural network for event-to-video conversion. We use the lightweight pretrained model with fixed-duration windows of 20\,ms, producing frames at 50\,Hz. Contrast-limited adaptive histogram equalization (CLAHE) and light Gaussian smoothing are applied to improve the low-contrast E2VID output. Dense optical flow is then computed between consecutive frames using OpenCV's Dense Inverse Search (DIS) algorithm, with the previous frame's flow used as initialization for temporal consistency.

The spatially-averaged flow direction over moving pixels (magnitude $>$ 0.3\,px) at each timestamp provides a per-frame ground-truth velocity sign, which we use to compute directional accuracy automatically across the full sequence rather than only within manually selected segments.

We note that this pipeline (events $\to$ E2VID $\to$ DIS optical flow) introduces its own errors: E2VID reconstructions are approximate, and DIS flow inherits any artifacts. We treat this as an approximate reference rather than exact ground truth, and cross-validate against the manual tracking to ensure consistency.

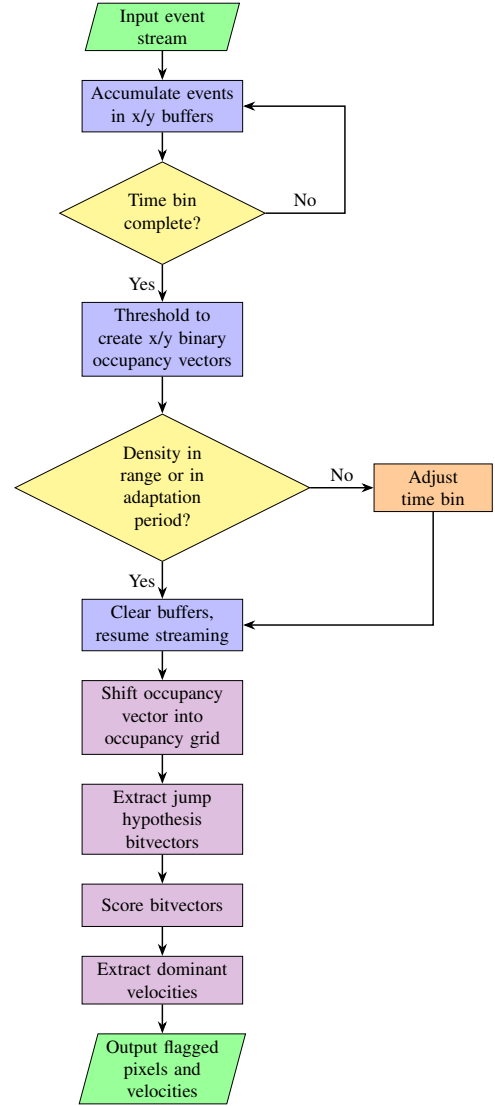
\begin{figure}
    \centering
    \scalebox{0.7}{  
            \begin{tikzpicture}[node distance=0.55cm]
    \node (input)     [io]                          {Input event stream};
    \node (accum)     [process, below=of input]     {Accumulate events in x/y buffers};
    \node (timebin)   [decision, below=of accum]    {Time bin complete?};
    \node (threshold) [process, below=0.7cm of timebin] {Threshold to create x/y binary occupancy vectors};
    \node (density)   [decision, below=0.7cm of threshold] {Density in range or in adaptation period?};
    \node (adjust)    [adjust, right=1.2cm of density]  {Adjust time bin};
    \node (clear)     [process, below=0.7cm of density] {Clear buffers, resume streaming};
    \node (shift)     [pipeline, below=of clear]    {Shift occupancy vector into occupancy grid};
    \node (extract)   [pipeline, below=of shift]    {Extract jump hypothesis bitvectors};
    \node (score)     [pipeline, below=of extract]  {Score bitvectors};
    \node (sort)      [pipeline, below=of score]    {Extract dominant velocities};
    \node (output)    [io, below=of sort]           {Output flagged pixels and velocities};
    
    \draw[arr] (input)     -- (accum);
    \draw[arr] (accum)     -- (timebin);
    \draw[arr] (timebin)   -- node[left]  {Yes} (threshold);
    \draw[arr] (threshold) -- (density);
    \draw[arr] (density)   -- node[left]  {Yes} (clear);
    \draw[arr] (clear)     -- (shift);
    \draw[arr] (shift)     -- (extract);
    \draw[arr] (extract)   -- (score);
    \draw[arr] (score)     -- (sort);
    \draw[arr] (sort)      -- (output);
    
    \draw[arr] (timebin.east)
        -- node[above] {No} ++(1.5,0)
        |- (accum.east);
    
    \draw[arr] (density.east) -- node[above] {No} (adjust.west);
    \draw[arr] (adjust.south) |- (clear.east);

        \end{tikzpicture}
    }
    \caption{Processing pipeline block diagram for the proposed event-based optical flow estimation. Incoming events 
    are accumulated into x/y buffers over adaptive time bins, thresholded into binary 
    occupancy vectors, and passed through a bitvector scoring stage to extract dominant 
    velocities.}
    \label{fig:pipeline}
\end{figure}

\section{Algorithm Description}

The proposed algorithm operates as a streaming pipeline: events are discretized into time bins, projected onto 1-bit occupancy vectors, stored in a shift register pipeline, and scored against discrete velocity hypotheses. The $x$ and $y$ axes are processed independently by identical pipelines, then combined. Fig.~\ref{fig:pipeline} shows the full datapath.

\subsection{Event Discretization}

An event camera produces an asynchronous stream of events $(t_k, x_k, y_k, p_k)$, where $t_k$ is the timestamp, $(x_k, y_k)$ is the pixel location, and $p_k \in \{-1, +1\}$ is the polarity. We discretize this stream into fixed-duration time bins of length $\Delta t$. For each bin $i$ spanning $[i\Delta t,\; (i+1)\Delta t)$, we count the number of events at each pixel location along the axis of interest.

For the $x$-pipeline, we collapse the $y$ coordinate and polarity, counting only how many events occur at each $x$-pixel regardless of $y$ or sign. A binary occupancy signal is then computed by thresholding:
\begin{equation}
E_x(x, i) = 
\begin{cases}
1, & \text{if } \sum_{k:\, x_k = x,\; t_k \in \text{bin } i} 1 \;\geq\; \theta_e \\
0, & \text{otherwise}
\end{cases}
\label{eq:occupancy}
\end{equation}
where $\theta_e$ is the event count threshold. The $y$-pipeline is identical, collapsing $x$ instead. This 1-bit quantization discards magnitude and polarity information but enables the entire downstream pipeline to operate on single-bit data.

The choice of $\Delta t$ and $\theta_e$ are coupled: a larger time bin accumulates more events per pixel, requiring a proportionally higher threshold to maintain sparse occupancy. We find that occupancy densities (fraction of active pixels per bin) in the 10--40\% range yield reliable results, with performance degrading below 10\% due to insufficient detections (see Section~\ref{sec: res}).

\subsection{Occupancy Grid}

The occupancy vectors from the $L$ most recent time bins are stored in a grid $G$ of size $N_x \times L$, where $N_x$ is the number of pixels along the axis and $L$ is the temporal depth (number of bins retained). Each entry $G[x, \ell]$ is a single bit: 1 if pixel $x$ was occupied at time bin $\ell$, 0 otherwise.

At each new time bin, the grid is updated by shifting all columns one position and inserting the new occupancy vector:
\begin{equation}
G[x, \ell] \leftarrow G[x, \ell+1] \quad \text{for } \ell = 0, \ldots, L-2
\label{eq:shift}
\end{equation}
\begin{equation}
G[x, L-1] \leftarrow E_x(x, i)
\label{eq:insert}
\end{equation}
In hardware, this is a bank of $N_x$ shift registers, each $L$ bits wide, updated in a single clock cycle via a parallel shift.

\subsection{Hypothesis Bitvector Extraction}
For each currently active pixel $x_0$ (where $E_x(x_0, i) = 1$), we evaluate a discrete set of velocity hypotheses $j \in \{-J, \ldots, +J\}$. Each hypothesis $j$ proposes that the feature at $x_0$ arrived there by moving $j$ pixels per time bin. To test this, we trace a diagonal path backward through the occupancy grid:
\begin{equation}
b_j(h) = G[x_0 - j \cdot h,\; L - h], \quad h = 1, \ldots, L
\label{eq:trace}
\end{equation}
where $h$ indexes how many time bins into the past we look. At each step, we check pixel $x_0 - j \cdot h$ at time offset $L - h$. If this pixel was occupied, it is consistent with a feature moving at velocity $j$ that now appears at $x_0$. The trace terminates early if $x_0 - j \cdot h$ falls outside $[0, N_x)$.

Two different hardware implementations of this search are considered. The most intuitive involves tracing these paths through the grid at each timestep, re-accessing the history at each new timestep. The second, lower-latency approach involves continuously collecting and comparing paths via spatial shift registers and clock enables, such that only the last two timesteps need to be considered to match the bitvector with the latest pixel associated with it.

\subsection{Hypothesis Scoring}
\label{sec:scoring}

We consider alternative scoring functions with different hardware cost profiles.

\subsubsection{Raw popcount} The simplest scorer counts the number of occupied cells along the trace:
\begin{equation}
R_j = \sum_{h=1}^{L} b_j(h)
\label{eq:popcount}
\end{equation}
Since $b_j(h) \in \{0,1\}$, the sum is a population count (popcount) of a binary vector --- implemented as a single up-counter that increments on each occupied cell. The counter width is $\lceil \log_2(L+1) \rceil = 5$~bits for $L=16$. Hypotheses with fewer than $\beta$ in-bounds steps are discarded ($R_j = 0$ if $H_j < \beta$), preventing high scores from short traces. To mitigate ties, when two hypotheses have the same raw score, the one with smaller $|j|$ is preferred, biasing toward slower motion hypotheses when evidence is ambiguous.

This scorer uses no DSP blocks, dividers, or floating-point units. The popcount operation is implemented with lookup-table (LUT)-based integer
logic, and the winner
is selected with fixed-width comparisons over  $b_s=\lceil \log_2(L+1)
\rceil$-bit score values.
However, large-$|j|$ hypotheses trace fewer in-bounds steps, so they are structurally disadvantaged. Setting $\beta \geq L/2$ ensures all surviving hypotheses have comparable trace lengths, making raw scores approximately comparable without normalization.

\subsubsection{Division-free normalized comparison} When hypotheses with widely varying in-bounds lengths must be compared, we avoid explicit division by cross-multiplying. To compare hypothesis $j$ against hypothesis $k$:
\begin{equation}
j \text{ beats } k \iff R_j \cdot H_k > R_k \cdot H_j
\label{eq:crossmul}
\end{equation}
Each product involves two values bounded by $L=16$, fitting in 8-bit arithmetic with no divider. 
Similarly, the threshold test $R_j / H_j > \theta_s / L$ becomes:
\begin{equation}
R_j \cdot L > \theta_s \cdot H_j
\label{eq:threshold}
\end{equation}
where multiplication by $L=16$ is a left shift by 4 bits, and $\theta_s \cdot H_j$ is a small constant-by-variable multiply (or a 16-entry lookup table indexed by $H_j$).

In the results reported in this paper, we use the raw popcount scorer with $\beta = L/2$ for synthetic data and the cross-multiplication comparison for real data, where trace lengths vary more due to objects near sensor edges.


\subsection{Winner Selection}

The dominant velocity at pixel $x_0$ is:
\begin{equation}
j^* = \arg\max_j \; S_j, \quad \text{subject to } S_{j^*} > \theta_s
\label{eq:winner}
\end{equation}
where $\theta_s$ is a score threshold above which a hypothesis is deemed worthy of notice. The physical velocity in pixels per second is:
\begin{equation}
v_x = \frac{j^*}{\Delta t}
\label{eq:velocity}
\end{equation}
In hardware, the $\arg\max$ is a comparator tree of depth $\lceil \log_2(2J+1) \rceil$ and the velocity is extracted from a lookup table to avoid divides.

\subsection{2D Extension}
\label{sec:2d}

The $x$ and $y$ axes are processed by independent, identical pipelines with separate occupancy grids $G_x$ ($N_x \times L$) and $G_y$ ($N_y \times L$). Each pipeline produces per-pixel velocity estimates along its axis: $v_x$ at each active $x$-pixel and $v_y$ at each active $y$-pixel.

To combine the two into a 2D velocity vector, we use the $y$-coordinates of events associated with each $x$-detection. When the $x$-pipeline produces a detection at pixel $x_0$, we record which $y$-pixels had events at $x_0$ during that bin. We then look up the $y$-pipeline's velocity estimate at those $y$-pixels and take the median as the associated $v_y$:
\begin{equation}
v_y(x_0) = \text{median}\{v_y(y) : y \in \mathcal{Y}(x_0)\}
\label{eq:vy_assoc}
\end{equation}
where $\mathcal{Y}(x_0)$ is the set of $y$-pixels with events at $x_0$ in the current bin. The output is a 2D velocity vector $(v_x, v_y)$ at each active $x$-pixel, along with the median $y$-coordinate for spatial localization.

This design doubles the hardware resources (two grids, two scoring pipelines) but introduces no new computational primitives. The two pipelines share no state during scoring and can execute fully in parallel, combining only at the final association step.

\subsection{Parameters}
\label{sec:params}

The algorithm has six parameters, summarized in Table~\ref{tab:params}. The most important design choice is $\Delta t$, which determines the displacement per bin and thus which velocities are resolvable. If $\Delta t$ is too small, the true displacement is sub-pixel and indistinguishable from noise; if too large, the occupancy grid saturates and all hypotheses score similarly.

\begin{table}[h]
\centering
\caption{Algorithm parameters and their roles.}
\label{tab:params}
\begin{tabular}{clc}
\toprule
Symbol & Description & Typical value \\
\midrule
$\Delta t$ & Time bin duration & 5--50\,ms \\
$\theta_e$ & Event count threshold & 30--100 \\
$L$ & Temporal depth (grid columns) & 16 \\
$J$ & Max hypothesis magnitude & 15 \\
$\beta$ & Min in-bounds steps & 4 \\
$\theta_s$ & Score threshold & $0.3L$ \\
\bottomrule
\end{tabular}
\end{table}

The hypothesis range $J$ is bounded by $\lfloor N_x / L \rfloor$: hypotheses with $|j| > N_x / L$ cannot stay in bounds for more than $L$ steps from any starting pixel. In practice, we set $J$ slightly below this bound. The event threshold $\theta_e$ scales approximately linearly with $\Delta t$, since longer bins accumulate proportionally more events.

%% file: Sections/hardware.tex
\section{Hardware Architecture}
\label{sec:hw}

This section presents an evaluation of the proposed algorithm in an FPGA platform, with the respective resource and latency estimations.

\subsection{Event Binning}

A free-running counter increments on each clock cycle and resets when it reaches $\Delta t / T_{\mathrm{clk}}$ cycles, issuing a bin-complete pulse. A small input FIFO decouples the asynchronous event stream from the synchronous scoring pipeline. During each bin, a bank of $N_x$ event counters (one per pixel, 8~bits each) accumulates incoming events, that are collapsed in the orthogonal direction. This allows us to use a buffer size that scales as $X+Y$ instead of $XY$, where $X$ and $Y$ define the camera dimensions. On the bin-complete pulse, each counter is compared against $\theta_e$; the result (1~bit per pixel) forms the new occupancy vector. All counters are then cleared for the next bin. No modulo logic is required — the bin timer is a simple compare-and-reset counter.

Adaptive bin duration using occupancy density 
feedback eliminates manual $\Delta t$ tuning, requiring only a popcount and two comparators. To allow for stabilization and easier velocity extraction, we propose an adjustment period during which the new timebin cannot be modified following each change.

\subsection{Occupancy Grid}

The occupancy grid is a bank of $N_x$ shift registers, each $L$~bits wide. On each bin-complete pulse, all registers shift left by one position in parallel, and the new occupancy bit is written into the rightmost position. This update takes a single clock cycle. No addressing logic or RAM controllers are needed — each shift register is a chain of flip-flops with a common clock enable.

\subsection{Hypothesis Scoring}

For each active pixel $x_0$, the scoring module evaluates $2J+1$ hypotheses $j \in \{-J, \ldots, +J\}$ in parallel. Each hypothesis lane $j$ contains:

\begin{itemize}
\item An index register initialized to $x_0$, decremented by $j$ each step
\item A bounds comparator checking $0 \leq x_{\mathrm{check}} < N_x$
\item A 5-bit up-counter accumulating occupied cells
\item A 4-bit step counter tracking $H_j$
\end{itemize}

All lanes trace simultaneously through the grid over $L$ clock cycles. Since all lanes read different $x$-addresses of the same grid in the same cycle, the grid must support $2J+1$ simultaneous reads. At $N_x = 240$ pixels $\times$ $L = 16$ bits $= 3{,}840$~bits, register-based storage is feasible and provides unlimited read ports. For larger sensors, banked SRAM with time-multiplexed access would be required.

After $L$ cycles, each lane holds a raw score $R_j$ and step count $H_j$. The winner is found by a pipelined comparator tree of depth $\lceil \log_2(2J+1) \rceil $ stages.

\subsection{Parallelism and Latency}

The architecture processes one active pixel at a time, with all hypothesis lanes executing in parallel. The latency per pixel is $L + \lceil \log_2(2J+1) \rceil$ clock cycles: $L=16$ cycles for the trace plus 5 cycles for the comparator tree, totaling 21 clock cycles per active pixel. At a 100~MHz clock, this is 210~ns per pixel.

If $n_a$ pixels are active in a given bin (typically 10--20\% of $N_x$), the total scoring time per bin is $21 \times n_a$ cycles. For the worst case of $n_a = N_x$ active pixels, the scoring completes in 5040 cycles or 50.4 \unit{\micro \second} at 100~MHz, which is well within a typical $\Delta t$ of 5--50~ms. The remaining bin time is idle, leaving substantial margin for lower clock frequencies or additional processing. 

\subsection{Incremental Scoring Variant}
\label{sec:incremental}
For the sensor and parameters considered in this work, the 
trace-based scorer completes well within a single time bin 
(under 1\% utilization at 100\,MHz). We describe an incremental variant for scenarios requiring shorter bins, larger sensors or smaller FPGAs with slower clock speeds. 

The trace-based scorer described above re-reads $L$ cells per hypothesis per active pixel, requiring $L$ clock cycles per pixel. We observe that between consecutive time bins, the diagonal trace for hypothesis $j$ at pixel $x_0$ is identical to the trace at pixel $x_0 + j$ in the previous bin, shifted by one timestep. The oldest occupancy value falls off one end and one new value enters at the other.

This permits an incremental update. We maintain a running score array $\hat{R}[x][j]$ of size $N_x \times (2J+1)$. On each bin-complete, the update for each pixel and hypothesis is:
\begin{equation}
\hat{R}[x_0][j] \leftarrow \hat{R}[x_0 + j][j] - G[x_0 - j \cdot L,\; 0] + G[x_0,\; L-1]
\label{eq:incremental}
\end{equation}
where $G[x_0 - j \cdot L, 0]$ is the oldest bit that falls off the trace and $G[x_0, L-1]$ is the new occupancy bit. This replaces the $L$-cycle trace with a single read--subtract--add--write per (pixel, hypothesis) pair, reducing the per-pixel scoring latency from $L$ to 1 clock cycle.

The cost is additional storage: $N_x \times (2J+1) \times \lceil\log_2(L+1)\rceil$ bits for the score array. For $N_x = 240$, $J = 15$, $L = 16$, this is $240 \times 31 \times 5 = 37{,}200$~bits ($\sim$4.5~kB) per axis. Table~\ref{tab:variant_comparison} compares the two approaches.

\begin{table}[h]
\centering
\caption{Trace-based vs.\ incremental scoring comparison (single axis, $N_x = 240$, $L = 16$, $J = 15$).}
\label{tab:variant_comparison}
\begin{tabular}{lcc}
\toprule
 & Trace-based & Incremental \\
\midrule
Cycles per pixel & $L + \log_2(2J+1) $ & $1 + \log_2(2J+1)$ \\
Total storage & $\sim$6 kbit & $\sim$43 kbit \\
Grid reads per bin & $L \times n_a \times (2J+1)$ & $n_a \times (1+2J+1)$ \\
\bottomrule
\end{tabular}
\end{table}

Both variants produce identical scores. The trace-based approach is simpler and uses less storage, making it attractive for initial implementation. The incremental variant trades storage for a $\sim$3.5$\times$ latency reduction (from 21 to 6 cycles per pixel), and would be preferred in applications requiring higher throughput or supporting larger sensors where the per-pixel trace cost becomes a bottleneck.

The results in this paper were validated using the trace-based approach. Verification of the incremental variant's equivalence and its FPGA synthesis are left to future work.

\subsection{Resource Estimates}

Table~\ref{tab:resources} summarizes the expected resource usage for the core datapath of our single-axis pipeline targeting a DAVIS240C sensor ($N_x = 240$, $L = 16$, $J = 15$). The values exclude board-facing I/O, buffering, debug, and implementation overhead.

\begin{table}[h]
\centering
\caption{Estimated core datapath storage for one axis of the EventShiftFlow pipeline
with $N_x=240$, $L=16$, and $J=15$.}
\label{tab:resources}
\begin{tabular}{lrc}
\toprule
Component & Bits / Units & Notes \\
\midrule
Occupancy grid & 3{,}840 bits & $240 \times 16 \times 1$-bit FFs \\
Event counters & 1{,}920 bits & $240 \times 8$-bit counters \\
Score accumulators & 155 bits & $31 \times 5$-bit (per lane) \\
Step counters & 124 bits & $31 \times 4$-bit (per lane) \\
Comparator tree & 5 stages & $\lceil \log_2(31) \rceil$ depth \\
\midrule
\textbf{Total (one axis)} & \textbf{$\sim$6{,}100 bits} & $< 1$~kB \\
\textbf{Total (both axes)} & \textbf{$\sim$13{,}000 bits} & $< 2$~kB \\
\bottomrule
\end{tabular}
\end{table}

For comparison, EDFLOW~\cite{liu2022edflow} uses 390 Block RAMs (855~kB) and 669 DSP48E units on a Xilinx Zynq XC7Z100, and Aung et al.~\cite{aung2018fpga} use 138 Block RAMs and 16 DSPs. Our design requires zero Block RAMs (all storage fits in distributed flip-flops), and zero DSP units.
The entire two-axis pipeline would occupy a small fraction of even low-cost FPGA devices.

\subsection{Division Avoidance}

The algorithm as described in Section~\ref{sec:scoring} avoids all division. The raw popcount scorer requires only a counter and comparators. The normalized variant uses cross-multiplication of 4-bit values (Eq.~\ref{eq:crossmul}), which can be implemented as a small combinational circuit or a 256-entry lookup table. The velocity output $v = j^* / \Delta t$ is computed off-chip by the host processor, since $\Delta t$ is a fixed configuration parameter. No arithmetic beyond counting and comparing occurs in the real-time datapath.

%% file: Sections/results.tex
\section{Results\label{sec: res}}
\begin{figure}[t]
\centerline{\includegraphics[width=0.999\columnwidth]{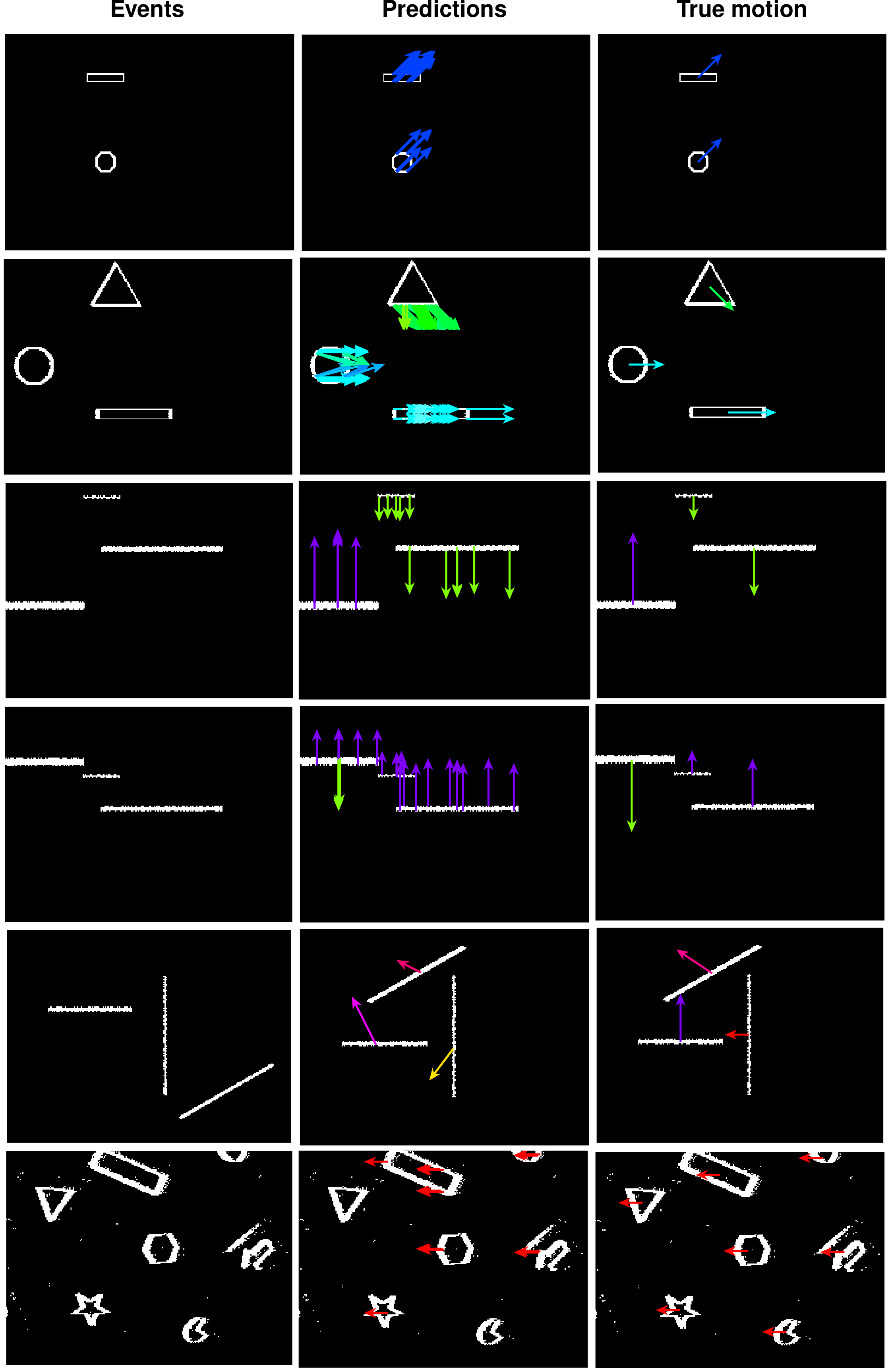}}
\caption{Event frames, predicted optical flow, and ground truth motion for six test scenarios. Each row shows a different scene: two synthetic single-object cases (rows 1–2), two synthetic multi-object cases (rows 3–4), a synthetic mixed-direction case (row 5), and a real-data scene (row 6). Arrow direction and colour encode the estimated velocity via the HSV colour wheel; arrow length is proportional to speed. The algorithm successfully estimates velocity direction and magnitude for the majority of detections across both synthetic and real data, with missed detections occurring primarily for objects with low event density or where objects overlap in the projected axis.}
\label{fig:frames}
\end{figure}

\subsection{Synthetic Data}

We first validate the algorithm on synthetic data with exact ground truth. The algorithm was validated in progressive stages: initially on a single slider moving horizontally, then on multiple sliders at different heights, on multiple bars with arbitrarily angled velocity vectors and finally on combinations of shapes moving at different angles. We also add 5\% additive gaussian noise events.

For uniaxial motion, near perfect accuracy in both magnitude and direction is observed aside from when bars travelling at different speeds overlap. The same phenomenon appears when multiple bars are travelling across the image at different angles as there are more overlap pixels. Figure~\ref{fig:frames} compares the algorithm's predicted velocities against the expected motion and illustrates several key behaviors of the proposed approach. Row 1 shows the algorithm's ability to detect diagonal motion not restricted to a single axis and row 2 presents the case of different shapes, where the circle and rectangle move horizontally whilst the triangle moves diagonally. We observe small perturbations in the predictions of the shapes whose paths are about to interfere, though the directionality is preserved. The same kind of behavior can be observed in rows 4 and 5. In section \ref{disc_sec}, we propose temporal regularization strategies to mitigate these effects.

\subsection{Real Data}

We evaluate on the \textit{shapes\_rotation} sequence from the RPG Event Camera Dataset~\cite{mueggler2017dataset}, using parameters $\Delta t = 40{,}000\,\mu$s, $\theta_e = 80$, $L = 16$, $J = 15$, $\beta = 4$, $\theta_s = 0.5L$.

\subsubsection{Directional accuracy}

Table~\ref{tab:dir_accuracy} reports per-segment directional accuracy, computed against manually tracked ground-truth velocity segments. The algorithm correctly identifies the dominant motion direction in all four segments, with high confidence in the leftward segments (negative $j^*$) and moderate confidence in the rightward segments.

\begin{table}[ht]
\centering
\caption{Per-segment directional accuracy on \textit{shapes\_rotation}.}
\label{tab:dir_accuracy}
\begin{tabular}{ccccc}
\toprule
Time (s) & GT avg. $j$ & Median $j_{\text{pred}}$ & Dir.\ acc.\ (\%) & $n$ \\
\midrule
0.78--1.10 & $2.50$  & $2$  & 94.7  & 19 \\
1.43--2.11 & $-2.50$ & $-3$ & 100.0 & 106 \\
2.32--3.11 & $3.00$  & $3$  & 99.4  & 174 \\
3.32--3.73 & $-3.46$ & $-4$ & 100.0 & 142 \\
\midrule
\multicolumn{2}{c}{Overall} & & 99.5\% & \\
\bottomrule
\end{tabular}
\end{table}


\subsubsection{Ground truth comparison}

\begin{figure}[t!]
\centerline{\includegraphics[width=0.95\columnwidth]{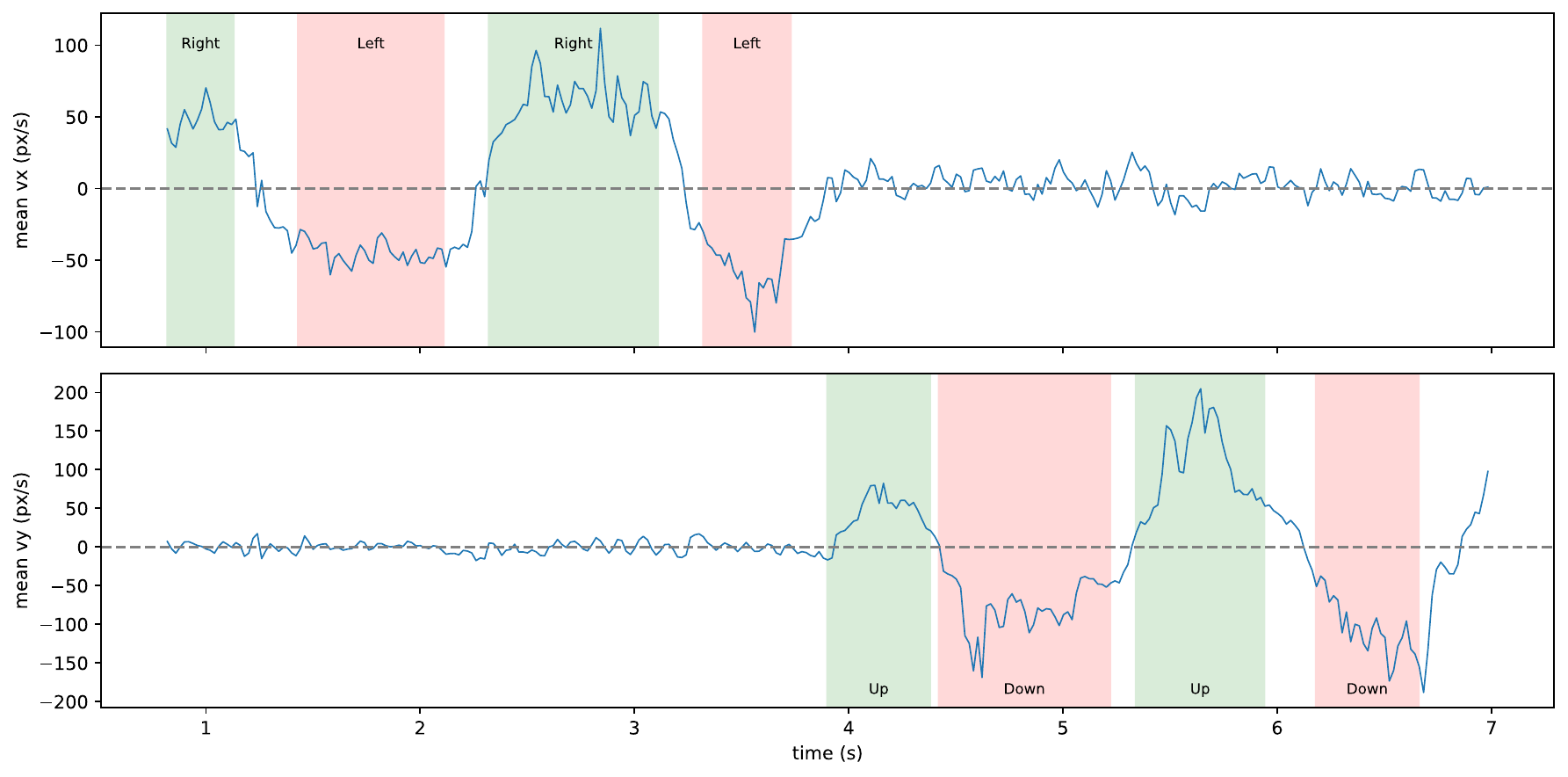}}
\caption{Predicted direction in red and green overlaid on E2Vid + CLAHE flow estimates. }
\label{fig: e2vid}
\end{figure}

Fig.~\ref{fig: e2vid} compares the algorithm's mean estimated direction and velocity against a frame-based ground-truth reference derived from E2VID~\cite{rebecq2019e2vid} reconstruction followed by DIS optical flow (Section~\ref{sec:gt}). The sign of the estimated velocity tracks the ground-truth sign across most of the sequence, with transitions between leftward and rightward motion occurring at approximately the correct timestamps. 

The last row of Figure \ref{fig:frames} compares the algorithm's predictions to the shapes' true motion, showcasing its ability to correctly predict the direction and magnitude ($\pm 1$px/$\Delta $t) of 5 out of 7 objects, especially the ones with flatter edges with respect to the direction of motion. These produce stronger projections onto the detection axis.



\subsubsection{Parameter sensitivity}

\begin{figure}[t]
\centerline{\includegraphics[width=0.999\columnwidth]{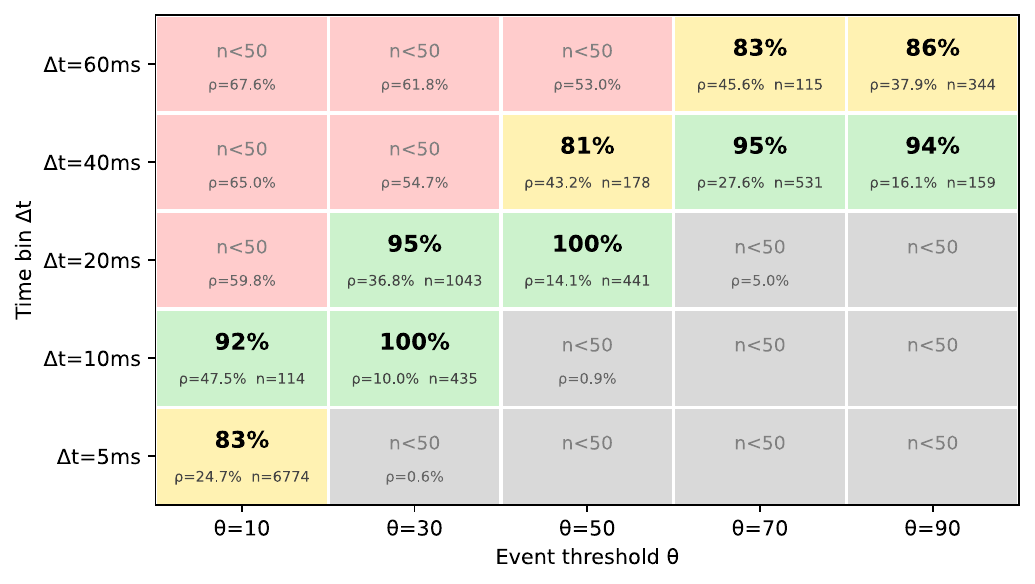}}
\caption{Directional accuracy across a grid of time-bin duration $\Delta t$ and event threshold $\theta$. Cell colour indicates accuracy where sufficient detections exist ($n \geq 50$): green ($\geq$90\%) and yellow (75--90\%). Red cells have high occupancy density ($\rho \geq 10$\%) but insufficient detections, while grey cells have low density ($\rho < 10$\%) and insufficient detections. Density $\rho$ and detection count $n$ are shown in each cell.}
\label{fig:density_sweep}
\end{figure}

Figure~\ref{fig:density_sweep} investigates how directional accuracy varies with fixed $\Delta t$ and $\theta_e$ on the \textit{shapes\_rotation} sequence. The two parameters are coupled: longer bins accumulate more events per pixel, requiring higher thresholds to maintain sparse occupancy. The best performance occurs when the resulting occupancy density falls in the 10--40\% range.


Setting $J$ too low truncates the detectable velocity range, causing the algorithm to miss fast-moving features. Setting $J$ too high wastes hypothesis lanes on velocities that exit the sensor bounds within a few steps. For the 240-pixel sensor with $L = 16$, $J = 15$ covers the full range of hypotheses that can stay in bounds for at least $\beta = 4$ steps.

\section{Preliminary FPGA Implementation}
\label{sec:preliminary-fpga}

To substantiate the hardware feasibility of EventShiftFlow, we implemented a
preliminary SystemVerilog prototype on a Xilinx Artix-7 FPGA. The prototype
accepts UART-encoded event words, parses them into 64-bit events, buffers the
stream through an AXI-stream-style FIFO, bins events into fixed-duration
temporal windows, thresholds each bin into a packed one-dimensional $x$-axis
occupancy vector, stores recent occupancy rows in a circular grid buffer, and
evaluates a discrete set of motion hypotheses over this grid. This prototype implements the one-axis scoring path used to validate the streaming hardware structure. To simplify timing closure and reduce implementation complexity, the current design evaluates motion hypotheses sequentially rather than using the fully parallel architecture described in Section V. Extending the same pipeline to the orthogonal axis and increasing hypothesis parallelism are left for future integration.

The scoring unit was implemented as a sequential pipeline rather than a fully
combinational search. For each completed occupancy bin, the scorer evaluates one
$(x,\Delta x)$ candidate at a time. The first stage computes the trajectory
score across the temporal grid and registers the candidate score. The second
stage compares the registered candidate against the score threshold and updates
the best motion estimate if the candidate improves on the current best. This
avoids placing the full $N_x \times N_h \times N_t$ search in a single clock
cycle while preserving deterministic execution time.

For the evaluated configuration, the design uses $N_x=240$ spatial bins,
$N_t=16$ temporal grid rows, and five displacement hypotheses
$\{-4,-2,0,2,4\}$. The scorer therefore requires at most
$2N_xN_h = 2400$ cycles per completed occupancy bin. At 100 MHz, this
corresponds to approximately $24~\mu$s of scoring latency.

Table~\ref{tab:fpga-results} summarizes the post-implementation results. The
reported utilization includes the complete board-facing prototype, including
UART receive/transmit framing and LED debug output, rather than only the
motion-scoring core. The design was synthesized and implemented in Vivado
2025.2 for a Xilinx Artix-7 \texttt{xc7a100tftg256-2} device using the
\texttt{Performance\_Explore} implementation strategy. The completed bitstream
meets the 100 MHz clock constraint with positive setup slack. The implementation
uses no DSP blocks and no block RAM tiles.

\begin{table}[t]
\centering
\caption{Preliminary FPGA implementation results for the EventShiftFlow
UART-to-motion prototype. Results are from Vivado post-implementation reports
for a Xilinx Artix-7 \texttt{xc7a100tftg256-2}.}
\label{tab:fpga-results}
\begin{tabular}{ll}
\toprule
\textbf{Item} & \textbf{Result} \\
\midrule
FPGA device & Xilinx Artix-7 \texttt{xc7a100tftg256-2} \\
Tool & Vivado 2025.2 \\
Implementation strategy & \texttt{Performance\_Explore} \\
Implemented axis & $x$ axis \\
Clock constraint & 100 MHz \\
Worst setup slack & $+0.030$ ns \\
Total negative slack & $0.000$ ns \\
Failing endpoints & 0 \\
On-chip power estimate & 0.142 W \\
\midrule
Spatial bins & 240 \\
Temporal grid rows & 16 \\
Motion hypotheses & $\{-4,-2,0,2,4\}$ \\
Scoring latency & 2400 cycles, $24~\mu$s at 100 MHz \\
\midrule
Slice LUTs & 13,326 \\
Slice registers & 5,517 \\
Block RAM tiles & 0 \\
DSP blocks & 0 \\
\bottomrule
\end{tabular}
\end{table}

These results support the central hardware claim of the method: the tested
configuration avoids dividers, floating-point units, DSP blocks,
and dense memory structures. Occupancy rows are stored as packed bit vectors,
and motion hypotheses are evaluated using fixed-width integer arithmetic,
comparisons, counters, and registers. The complete prototype occupies
approximately 21\% of the available LUTs and 4\% of the available flip-flops on
the target Artix-7 device. These figures correspond to the complete UART-to-motion prototype, including UART communication, buffering, control logic and LED debug circuitry. The motion-scoring datapath itself occupies only a subset of the reported resources.

The prototype has been verified using cocotb and Verilator testbenches covering
UART event parsing, FIFO backpressure, temporal binning, occupancy thresholding,
multi-bin operation, occupancy-grid updates, and scorer behavior on positive,
stationary, and no-match trajectories. The current prototype assumes that the
occupancy grid is not overwritten during the scorer's short execution window;
future versions can remove this assumption using explicit backpressure or a
ping-pong grid buffer. Broader algorithmic evaluation on dynamic backgrounds
and standard optical-flow error metrics remains separate from this preliminary
FPGA validation.

%% file: Sections/outlook.tex
\section{Discussion\label{disc_sec}}




\subsection{Limitations}

The independent 1D pipeline architecture inherits the aperture problem: the $x$-pipeline cannot distinguish a bar moving purely rightward from one moving diagonally. The $y$-association step (Eq.~\ref{eq:vy_assoc}) partially mitigates this but fails when multiple objects at the same $x$-pixel move in different $y$-directions.

The discrete hypothesis set imposes a velocity resolution of $1/\Delta t$\,px/s. Objects moving at velocities between hypotheses are assigned to the nearest integer jump, producing quantization error that grows with $\Delta t$. Sub-pixel interpolation (e.g., parabolic peak fitting across adjacent hypotheses) could reduce this error but would require multipliers.

Parameter selection currently requires manual tuning. The coupling between $\Delta t$ and $\theta_e$ means that a single parameter set does not generalize across scenes with different event rates or velocity distributions.

\subsection{Comparison to Existing Methods}
The presented method uses little resources. The entire two-axis pipeline fits in under 13\,kbit of storage with no dividers and no floating-point units, compared to EDFLOW's 855\,kB of Block RAM and 669 DSP units. 
While our method does not produce optical flow as accurately as EDFLOW~\cite{liu2022edflow} or E-RAFT~\cite{gehrig2021eraft}, 
for applications such as reactive obstacle avoidance on micro aerial vehicles, knowing the dominant motion direction at low latency may be more valuable than precise flow vectors at higher cost and latency. The algorithm's streaming nature and fixed per-bin latency make it suitable as a first-stage motion detector that could trigger more expensive processing only when motion is detected.

\subsection{Future Work}


Validation against a live event camera and extension of the current prototype to the full two-axis architecture are immediate next steps. The present implementation demonstrates event ingestion, occupancy generation and one-axis motion scoring on FPGA; future work will continue optimizing the implementation for different hypotheses and sensor resolutions to then characterize scalability across devices and quantify performance under live sensing conditions.

The current method assumes that the dominant motion within a local region produces the strongest spatiotemporal correlation. In scenes containing multiple independently moving objects or highly dynamic backgrounds, competing event trajectories may generate similar correlation scores, reducing the ability to identify a unique dominant motion hypothesis. Additional experiments on more diverse datasets are therefore needed to better characterize robustness and identify the operating regimes in which the method performs reliably. Comparison against established event-based optical flow methods such as EV-FlowNet\cite{zhu2018evflownet} would additionally enable evaluation using standard optical-flow metrics, including endpoint error.

Several algorithmic extensions are also of interest. Temporal regularization, such as majority voting across recent bins, exponential smoothing of the dominant hypothesis, or a score bonus for the previous winner, may improve robustness to single-bin fluctuations at minimal hardware cost. Multi-bit occupancy representations (e.g., 4-bit event counts rather than binary occupancy) could provide additional scoring discrimination at the expense of increased memory requirements, while polarity-aware scoring could exploit the sign consistency of moving edges. Finally, the incremental scoring variant presented in Section~\ref{sec:incremental} may enable shorter time bins or operation on higher-resolution sensors.

\section{Conclusion}
We presented EventShiftFlow, a streaming velocity estimator for event cameras that, in its minimal configuration, operates using only shift registers, counters, and comparators. The algorithm discretizes events into time bins, stores 1-bit occupancy in a shift-register grid, and evaluates discrete velocity hypotheses by counting coincidences along diagonal traces. On synthetic data the method achieves near-perfect directional accuracy; on real event camera data it correctly identifies motion direction in all four evaluated segments with an overall directional accuracy of 99.5\%, with performance remaining robust across occupancy densities in the 10--40\% range. The entire two-axis pipeline requires under 13\,kbit of on-chip storage and no dividers, no floating-point units, and no DSP blocks, representing several orders of magnitude lower memory usage than EDFLOW.